\documentclass[manuscript,nonacm]{acmart}

\usepackage{comment}

\usepackage{enumitem}

\newcommand{\bisinfo}{{bisinformation}} 
\newcommand{\misinfo}{{misinformation}} 

\setlist{itemsep=0pt,leftmargin=*}

\usepackage{titlesec}

\titlespacing*\section{0pt}{3pt plus 2pt minus 2pt}{3pt plus 2pt minus 2pt}
\titlespacing*\subsection{0pt}{3pt plus 2pt minus 2pt}{3pt plus 2pt minus 2pt}
\titlespacing*\subsubsection{0pt}{3pt plus 2pt minus 2pt}{3pt plus 2pt minus 2pt}
\titlespacing\paragraph{0pt}{3pt plus 2pt minus 2pt}{3pt plus 2pt minus 2pt}

\titleformat{\section}{\bfseries\large}{\thesection}{1em}{}
\titleformat{\subsection}{\bfseries\large}{\thesubsection}{1em}{}
\titleformat{\subsubsection}{\bfseries\normalsize}{\thesubsubsection}{1em}{}
\titleformat{\paragraph}{\bfseries\normalsize}{}{1em}{}

\AtBeginDocument{%
  \providecommand\BibTeX{{%
    \normalfont B\kern-0.5em{\scshape i\kern-0.25em b}\kern-0.8em\TeX}}}

\setcopyright{none}
\copyrightyear{}
\acmYear{}
\acmDOI{}

\begin{document}

\title{Fairness via AI: Bias Reduction in Medical Information}

\author{Shiri Dori-Hacohen}
\email{shiridh@uconn.edu}
\orcid{1234-5678-9012}
\affiliation{%
  \institution{University of Connecticut \& AuCoDe}
  \country{USA}
}

 \author{Roberto Montenegro}
\affiliation{%
  \institution{Seattle Children's Hospital}
  \country{USA}
}

\author{Fabricio Murai}
\affiliation{%
  \institution{Universidade Federal de Minas Gerais}
  \country{Brazil}}

 \author{Scott A.\ Hale}
\affiliation{%
  \institution{Meedan}
  \country{USA}
  }

\author{Keen Sung}
\affiliation{%
  \institution{AuCoDe}
  \country{USA}
  }

\author{Michela Blain}
\affiliation{%
  \institution{University of Washington School of Medicine}
  \country{USA}
}

\author{Jennifer Edwards-Johnson}
\affiliation{%
  \institution{Michigan State University College of Human Medicine}
  \country{USA}
}

\renewcommand{\shortauthors}{Dori-Hacohen, et al.}


\maketitle

\section{Introduction} 

Most Fairness in AI research focuses on exposing biases in AI systems. A broader lens on fairness reveals that AI can serve a greater aspiration: rooting out societal inequities from their source. Specifically, we focus on inequities in health information, and aim to reduce bias in that domain using AI. The AI algorithms under the hood of search engines and social media, many of which are based on recommender systems, have an outsized impact on the quality of medical and health information online. Therefore, embedding bias detection and reduction into these recommender systems serving up medical and health content online could have an outsized positive impact on patient outcomes and wellbeing. 

In this position paper, we offer the following contributions: 
(1) we propose a novel framework of Fairness \textbf{via} AI, inspired by insights from medical education, sociology and antiracism; 
(2) we define a new term, \textbf{bisinformation}, which is related to, but distinct from, misinformation, and encourage researchers to study it; 
(3) we propose using AI to study, detect and mitigate biased, harmful, and/or false health information that disproportionately hurts minority groups in society; 
and
(4) we suggest several pillars and pose several open problems in order to seed inquiry in this new space.
While part (3) of this work specifically focuses on the health domain, the fundamental computer science advances and contributions stemming from research efforts in bias reduction and Fairness via AI have broad implications in all areas of society.

\section{Fairness via AI}

The vast majority of Fairness in AI work focuses on exposing the bias in AI system, in order to showcase where the AI system is biased. However, AI systems will continue to be biased so long as the data they are receiving are biased, according to ``Bias In, Bias Out'' principle \cite{mayson2018bias}. So long as significant structural inequalities exist in the real world, AI systems will continue to replicate and exacerbate them. The dominant Fairness in AI approach, then, risks engaging in a Sisyphean task of minimizing bias in AI, with attempts to debias AI datasets, models and algorithms continually needing to be ``fixed'' as they learn biased outcomes and are bound to hit a ceiling of fairness: that of real world settings that are inherently biased. Under this approach, our highest aspiration in designing AI systems seems to be one of avoidance: tweaking our models to refrain from adding to society's ills and inequities. 

Rooted in insights from medical education, sociology, and antiracism, we offer a broader lens on fairness, revealing that AI can serve a far greater aspiration: enabling important restorative work and rooting out societal inequities from their source, with a deeper and more meaningful impact. In this position paper, we therefore reverse the traditional direction of fairness: rather than aiming to achieve Fairness \textbf{in} or \textbf{of} AI, we propose focusing on \textbf{Fairness via AI}. With this approach, one can use AI to study, detect, mitigate and remedy situations that are inherently unequal, unjust and unfair in society. With this ambitious yet grounded approach, our potential impact is unbounded, and can accelerate progress towards a more fair, equal and just world. In other words, we can use AI to thoughtfully, carefully and ethically debias the world, rather than simply trying to debias AI.
Specifically, the AI algorithms under the hood of search engines and social media, many of which are based on recommender systems, have an outsized impact on the quality of information available online. Therefore, embedding bias detection and reduction directly into these recommender systems could have an outsized positive impact on the information ecosystem. 


\section{Bisinformation}

We coin the term \textbf{\bisinfo{}} to represent biased information, referring to a unique and challenging aspect of the information landscape. We are particularly interested in health bisinformation,  where bias and language misuse have a detrimental impact on patient outcomes, though the term can easily apply in any field. Bisinformation 
may overlap with, but is not identical to, misinformation. The use of biased language or inappropriate social identifiers in a medical context, for example, can be harmful even if strictly true - consider the case of referring to the prevalence of an illness in a racial category without contextualizing it in Social or Structural Determinants of Health (SSDoH), such as systemic racism or income inequities~\cite{metzl2019structural}. On the other hand, certain types of \bisinfo{} are, in fact, also a form of misinformation, such as the long-debunked Salt Gene Hypothesis~\cite{pollock20124}.

To the best of our knowledge, no studies have computationally studied health \bisinfo{} at a large scale.

\textbf{Medical \bisinfo{} (and misinformation).} As an illustration of societal inequities in dire need of the Fairness via AI approach, consider the field of medicine and medical education. The field is marred by a long and painful history of overt and covert forms of social injustice, bias, and racism, as illustrated by the American Medical Association’s recent pledge to take action to confront systemic racism~\cite{saini2019superior, madara2020}. Studies continue to demonstrate that physicians possess implicit biases in a number of different areas such as race/ethnicity, gender, sex, age, weight, substance use and mental illness \cite{fitzgerald2017implicit}. This comes into play significantly in medical institutions, which continue to teach biased medicine in preclinical years \cite[see, e.g.][]{tsai2016race}. Many educators, for example, continue to inappropriately use race as a proxy for genetics or ancestry, or even as a ``risk factor'' for numerous health outcomes often erroneously associated with race while ignoring SSDoH~\cite{ali2011use, hunt2013genes, acquaviva2010perspective, karani2017commentary, metzl2019structural}. Many educators continue to inappropriately use gender and sex terms and perpetuate the idea that sex and gender are binary and stagnant (versus fluid). Likewise, most medical educators are unaware of the numerous biases in the types of images they use in their lectures or assessment materials as well~\cite{doja2016hidden, fallin2015implicit, lempp2004hidden}. 
By equating social identifiers to biology without social or structural context, medical educators are unknowingly perpetuating a curriculum that can have an adverse effect on health outcomes \cite{martimianakis2015humanism, braun2007racial}. Bias reduction in curricular and assessment content is critical for educating future physicians in accurate evidence-based medicine \cite{le2018first, ripp2017race}, but is a manual, costly and time consuming effort. SOTA AI and NLP approaches can be used to scale up these efforts significantly\footnote{Montenegro, Dori-Hacohen and Sung have recently been awarded a grant from the National Board of Medical Examiners' Stemmler Fund to reduce bias in medical curricular content using NLP and ML approaches.}. 

Naturally, \bisinfo{} and misinformation that persist in the medical establishment, are also disseminated and extensively present in online medical resources, websites and news articles, and social media, with large negative effects on historically underserved populations, also reinforcing biased narratives and stereotypes about minority groups. Recommender systems play an outsized role in serving up such content online, but improving these systems to reduce bias will prove extremely challenging if we don't understand the underlying mechanisms in which such bias is perpetuated and disseminated. Prior work has suggested that controversy online is highly unevenly distributed, and that a population-sensitive model is needed in order to properly model this~\cite{jang2017modeling}; we hypothesize that the same approach may be needed in the computational study of bisinformation and misinformation.
For example, the COVID-19 pandemic and its associated ``infodemic'' has brought health mis- and disinformation to the forefront of national and scientific attention. However, trust in the medical establishment may be understandably low among African-Americans \cite{brandon2005legacy} and other minority groups~\cite{jaiswal2020disinformation}. Recent work suggests, moreover, that health mis- and disinformation is qualitatively distinct in different population groups \cite[e.g.][]{prasad2021,jaiswal2020disinformation}. To cite just one example, COVID-19 vaccine hesitancy has been demonstrated to be higher among racial and ethnic minorities \cite{Nguyen2021.02.25.21252402, jaiswal2020disinformation}.

\section{Pillars \& Open Problems}

A few guiding pillars underlie and drive our \textit{Fairness via AI} research agenda, which we encourage others to adopt. First, we argue that Fairness via AI is a more effective and impactful marshalling of research resources than ``standard'' Fairness in AI work (important though the latter may be). Second, we argue that collaboration across disciplinary fields is critically needed in order to effectively and ethically study and understand society's biggest challenges, to say nothing of mitigating them. Researchers in other fields, including but in no way not limited to the social sciences, have immense expertise in studying and addressing societal inequities; computer scientists cannot, and should not, go this alone. Finally, we argue that biased information interacts with false information in complex ways that must be studied carefully in order to reduce bias in recommender systems and other information delivery systems (such as search engines). 

With these pillars firmly in mind, we pose the following open questions: 

\begin{enumerate}[label=Q\arabic*.,itemindent=*,nosep]
    \item What are effective approaches to identify societal problems that are in most need of, and lend themselves to, the \textbf{Fairness via AI} framework?
    \item Which existing and/or novel AI approaches need to be deployed and developed in order to address such societal issues?
    \item How can we encourage collaboration across disciplinary boundaries in order to leverage hard-won insights from other fields, and infuse our Fairness research with them?
    \item Specifically with respect to \bisinfo{}, several research questions arise:
\begin{enumerate}[label=\alph*.,itemindent=*,nosep]
    \item  How and where does \bisinfo{} spread online? Is information (including mis- and \bisinfo{}) distributed and disseminated differentially among diverse population groups? If so, how? 
    \item Which categories or types of \bisinfo{} and misinformation are most problematic and harmful, and thus deserving the most diligent fact checking, and countermessaging efforts? In other words, how should we triage mis- and \bisinfo{}, combining best practices in the public health and fact checking spheres with state-of-the-art computational approaches?
\end{enumerate}
\end{enumerate}

\section{Conclusions}

In this position paper, we introduced a novel \textbf{Fairness via AI} framework; coined a new term, \bisinfo{}, to describe biased information, and demonstrated that is is overlapping with yet distinct from \misinfo{}; briefly discussed the documented presence of \bisinfo{} in medical curricula and posit that this extends to other information environments, such as online; and posed several open questions to guide research agendas on the subject.

\noindent\small{\textbf{\emph{Acknowledgements}}.
This material is based in part upon work supported by the National Science
Foundation under Grant No.~1951091. Any opinions, findings and conclusions or recommendations expressed in this material are those of the author(s) and do not necessarily reflect the
views of the National Science Foundation.}

\bibliographystyle{ACM-Reference-Format}
\bibliography{refs}

\end{document}